\documentclass[sigconf,natbib=false]{acmart}
\usepackage{colortbl}
\usepackage{xcolor}
\usepackage{soul} 
\usepackage{multirow}
\usepackage{boxedminipage} 
\usepackage{graphicx}
\usepackage{hyperref}
\usepackage{tabularx}
\usepackage{subcaption}
\usepackage[normalem]{ulem}
\usepackage{makecell}
\usepackage{csquotes}
\definecolor{gred}{RGB}{219,68,55}
\definecolor{gblue}{RGB}{66,133,244}
\definecolor{gyellow}{RGB}{244,180,0}
\definecolor{ggreen}{RGB}{15,157,88}
\definecolor{ggrey}{RGB}{115,115,115}

\definecolor{bleudefrance}{rgb}{0.19, 0.55, 0.91}
\definecolor{yes}{RGB}{239,211,69}
\definecolor{carminered}{rgb}{1.0, 0.0, 0.22}
\definecolor{crimsonglory}{rgb}{0.75, 0.0, 0.2}

\AtBeginDocument{%
  \providecommand\BibTeX{{%
    \normalfont B\kern-0.5em{\scshape i\kern-0.25em b}\kern-0.8em\TeX}}}

\copyrightyear{2024}
\acmYear{2024}
\setcopyright{rightsretained}
\acmConference[SIGIR '24]{Proceedings of the 47th International ACM SIGIR Conference on Research and Development in Information Retrieval}{July 14--18, 2024}{Washington, DC, USA}
\acmBooktitle{Proceedings of the 47th International ACM SIGIR Conference on Research and Development in Information Retrieval (SIGIR '24), July 14--18, 2024, Washington, DC, USA}
\acmDOI{10.1145/3626772.3657789}
\acmISBN{979-8-4007-0431-4/24/07}

\makeatletter
\patchcmd{\maketitle}{\@copyrightpermission}{
   \begin{minipage}{0.3\columnwidth}
     \href{http://creativecommons.org/licenses/by/4.0/}{\includegraphics[width=0.70\textwidth]{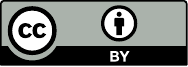}}
   \end{minipage}\hfill
   \begin{minipage}{0.7\columnwidth}
     \href{http://creativecommons.org/licenses/by/4.0/}{This work is licensed under a Creative Commons Attribution International 4.0 License.}
   \end{minipage}
}{}{}
\makeatother

\RequirePackage[
  datamodel=acmdatamodel,
  style=acmnumeric,
  ]{biblatex}

\begin{document}

\title{Flexible and Adaptable Summarization via Expertise Separation}

\author{Xiuying Chen}
\affiliation{%
  \institution{ KAUST, CBRC\\
  MBZUAI}
  \country{Jeddah, Saudi Arabia}
}
\email{xiuying.chen@kaust.edu.sa}

\author{Mingzhe Li}
\affiliation{%
  \institution{Ant Group}
  \country{Beijing, China}
}
\email{limingzhe.lmz@antgroup.com}

\author{Shen Gao}
\affiliation{%
  \institution{ Shandong University}
  \country{Shandong, China}
}
\email{shengao@sdu.edu.cn}

\author{Xin Cheng}
\affiliation{%
  \institution{ Peking University}
  \country{Beijing, China}
}
\email{chengxin1998@stu.pku.edu.cn}

\author{Qingqing Zhu}
\affiliation{%
  \institution{Peking University}
  \country{Beijing, China}
}
\email{zhuqingqing@pku.edu.cn}

\author{Rui Yan}
\affiliation{%
  \institution{Renmin University of China}
  \country{Beijing, China}
}
\email{ruiyan@ruc.edu.cn}

\author{ Xin Gao$^{\dagger}$}
\affiliation{%
  \institution{KAUST, CBRC}
  \country{Jeddah, Saudi Arabia}
}
\email{xin.gao@kaust.edu.sa}

\author{Xiangliang Zhang$^{\dagger}$}
\affiliation{%
  \institution{University of Notre Dame
  }
  \country{South Bend, USA}
}
\email{xzhang33@nd.edu}

\thanks{$\dagger$ Corresponding authors. Dr. Zhang is secondly affiliated with KAUST.} 

\renewcommand{\shortauthors}{Xiuying Chen et al.}

\begin{abstract}
A proficient summarization model should exhibit both \emph{flexibility }-- the capacity to handle a range of in-domain summarization tasks, and \emph{adaptability} -- the competence to acquire new knowledge and adjust to unseen out-of-domain tasks.
Unlike large language models (LLMs) that achieve this through parameter scaling,   we propose  a more parameter-efficient approach in this study.
Our motivation rests on the principle that the general summarization ability to capture salient information can be shared across different tasks, while the domain-specific summarization abilities need to be distinct and tailored.
Concretely, we propose MoeSumm, a Mixture-of-Expert Summarization architecture, which utilizes a main expert for gaining the general summarization capability and deputy experts that selectively collaborate to meet specific summarization task requirements.
We further propose a max-margin loss to stimulate the separation of these abilities.
Our model's distinct separation of general and domain-specific summarization abilities grants it with notable flexibility and adaptability, all while maintaining parameter efficiency.   
MoeSumm achieves \emph{flexibility} by managing summarization across multiple domains with a single model,  utilizing a shared main expert and selected deputy experts. 
It exhibits \emph{adaptability} by tailoring deputy experts to cater to out-of-domain few-shot and zero-shot scenarios.
Experimental results on 11 datasets show the superiority of our model compared with recent baselines and LLMs.
We also provide statistical and visual evidence of  the distinct separation of the two abilities in MoeSumm\footnote{\url{https://github.com/iriscxy/MoE_Summ}}.
\end{abstract}




\maketitle

\section{Introduction}

\begin{figure}
\centering
\includegraphics[scale=0.61]{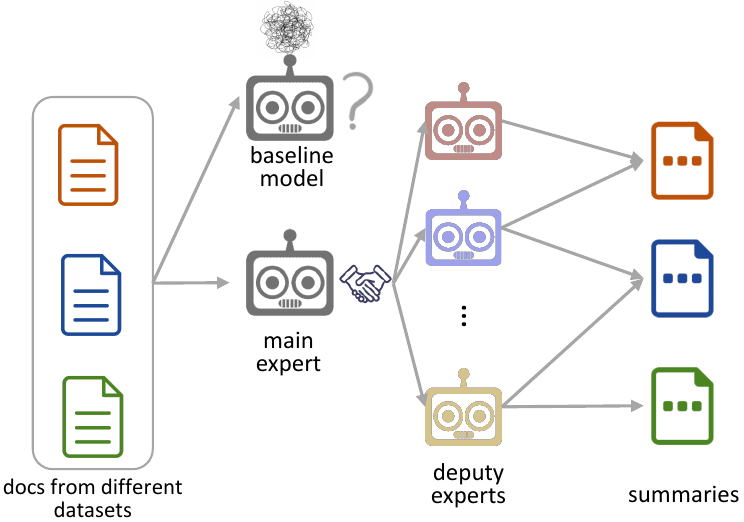}
\caption{
 Comparison of the existing summarization model and our MoeSumm model.
 Our MoeSumm consists of a main summarization expert and multiple deputy experts, which can be used for or quickly adapt to different datasets.
}
\label{fig:intro}
\vspace{-3mm}
\end{figure}

Text generation has made impressive progress in recent years~\cite{ouyang2022training,zhang2020summarizing,park2022qsg}.
The task of abstractive summarization, aiming to produce a concise, fluent, and faithful summary, has become a research hotspot due to its broad application prospect~\cite{wang2022ofa,chentowards}.
Herein, we outline two key capabilities that an intelligent summarization system should possess. 
The first is \textit{flexibility}, indicating the system's competence to be readily applied to a variety of in-domain  tasks. 
A  flexible summarization system should be proficient in summarizing various types of content, such as news articles and scientific papers.
The second ability is \textit{adaptability}, for acquiring new knowledge and adapting to unseen out-of-domain summarization tasks.
For example, a medical summarization model trained prior to 2019 needs to be able to adapt and acquire knowledge about COVID-19.

Existing pretrained summarization models typically use  a \textit{one-model-for-one-domain} approach, training separate models on individual datasets each optimized for a specific domain~\cite{See2017GetTT,li2021prefix}. However, this strategy hampers their flexibility as a model tailored for one domain may underperform in others~\cite{fu2023specializing}. Alternatively, recent LLMs like GPT-3~\cite{brown2020language} and GPT-3.5 exhibit remarkable summarization performance, powered by vast data volumes and computational resources~\cite{yang2023exploring}. However, these \textit{one-large-model-for-all-domains} frameworks have their drawbacks, including being closed-source, costly, and susceptible to data leakage~\cite{tian2023opportunities}.
Furthermore, their inability to edit or scale the knowledge embedded within them once trained results in limited adaptability to fresh knowledge~\cite{cheng2023decouple}.

Different from previous works,  this paper aims to improve summarization flexibility and adaptation in a parameter-efficient way, resulting in a \textit{one-small-model-for-all-domains} approach.
Our motivation  stems from the need for a general summarization capability to distill key input information and a specialized adaptability to refine this information in line with specific summarization requirements such as language style, summary length, and conciseness.
By sharing the general ability, a summarization model avoids redundant learning for each domain and focuses on common features, while separating specialized abilities ensures tailored, high-quality summaries without unnecessary complexity. 
Correspondingly, we propose a Mixture-of-Expert Summarization (\textit{MoeSumm}) model, where a main expert captures salient information, and deputy experts work with the main expert to adapt the extracted summary information to the different domains.
Specifically, we choose to incorporate expert separation by adapting the feed-forward neural networks (FFNs) in pretrained models.
The main expert collaborates with selected deputy experts to form an FFN. 
These deputy experts,  chosen by a dataset-aware gating function, are designed to learn dataset-aware summarization abilities. 
To prevent the model from over-relying on the main expert and collapsing into a single model, we propose a max-margin loss, where the margin is defined as the prediction difference brought by the deputy experts.
The max-margin loss is specifically designed to distinguish between different abilities without compromising performance.
Due to its decouple attribute, MoeSumm can naturally adapt to out-of-domain few-shot domains, where only the deputy experts need to be fine-tuned.
MoeSumm can also be used in zero-shot settings, where we can utilize the main expert to give a general summary.
For the few-short setting, we can use the few cases to finetune a new deputy expert, which can work with other well-learned modules to produce a suitable summary.

We validate the effectiveness of MoeSumm in 3 settings (in-domain, out-of-domain few-shot, and zero-shot) across 11 benchmark datasets.
The datasets are from various domains (news, academic papers, social media posts, etc.), varying in input and output lengths, levels of abstractiveness, and language style.
Experiment results show that our MoeSumm outperforms all baseline models in most of the metrics.
In addition, we demonstrate the separation of general and different specialization abilities through comprehensive experiments, which also provide an explanation for the generation.

\textcolor{black}{ 
Overall, our main contributions are:}
(1) We propose MoeSumm, a parameter-efficient summarization model that is  applicable to a variety of in-domain summarization tasks and is well-suited for out-of-domain few-shot and zero-shot summarization scenarios.
The model's parameter efficiency is ensured by the shared general summarization ability.
(2) To achieve the above goal, we design MoeSumm with a mixture-of-expert structure and a max-margin loss to distinguish the roles of the different experts.
(3) Experimental results demonstrate that   MoeSumm brings substantial improvements over strong baselines in  both in-domain and out-of-domain scenarios across benchmark datasets.

\section{Related Work} 
We discuss related work on one-model-for-one-domain and one-model-for-all-domain summarization, mixture-of-experts model, and lightweight fine-tuning.

\textcolor{black}{\textbf{Summarization.}}
\textcolor{black}{Most existing summarization models adopt a \textit{one-model-for-one-domain} approach.
For example, \cite{yu2021adaptsum,chen2022spec}  finetune all the parameters in the model for each target dataset.
\cite{fonseca2022factorizing} first conduct content selection and then output the summary.}
\cite{li2021prefix} proposed a lightweight finetuning method by adjusting a task-specific vector prefixed to the original inputs. 
While these models achieve good performance on specific summarization tasks, they tend to struggle in maintaining consistent effectiveness across a variety of domains~\cite{wang2019exploring}.
Moreover, as demonstrated by Fu et al.~\cite{fu2023specializing}, increasing specialized ability comes at the cost of decreased generic ability.
Another line of research aims to propose \textit{one-model-for-all-domains}.
Towards this goal, \cite{hua2017pilot} showed that a source dataset can help summarize the target data when it captures the style for a target domain.
\cite{wang2019exploring} 
developed a multi-domain extractive summarization model and examined the impact of domain discrepancies on extractive summarization performance. 
Despite demonstrating flexibility, the model didn't effectively address the adaptability challenge associated with out-of-domain tasks. 
\textcolor{black}{Furthermore, our focus on abstractive summarization presents a more substantial challenge due to its generative nature.}
Most recently, large language models such as GPT-3 have shown impressive flexible summarization abilities, made possible by vast data and computational resources. 
In contrast, 
we aim for a parameter-efficient approach that addresses both flexibility and adaptability.

\begin{figure*}[t]
\centering
\includegraphics[scale=0.52]{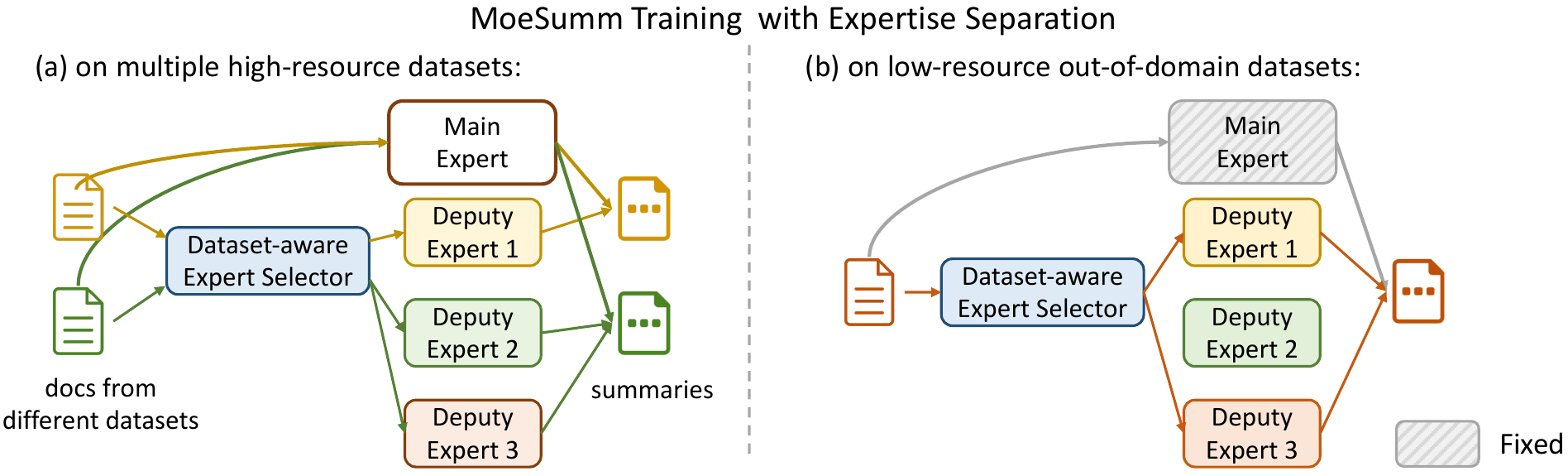}
\caption{ \textcolor{black}{Training MoeSumm under different settings. (a) Training the expert selector and all experts on multiple high-resource datasets. (b) Fine-tuning only the expert selector and the deputy experts on   low-resource datasets.  }
} 
\label{fig:frame}
\end{figure*}

\textbf{Mixture-of-Experts Models.}
MoE models were initially proposed to increase the model's capacity while maintaining a constant computational cost during inference, where a fixed number of experts are adaptively activated by an input during training and inference \cite{zuo2021taming,mustafa2022multimodal}.
Typically, a trainable gate in MoE determines the activation of experts, often resulting in an imbalance with most inputs routed to a single expert.
Various restriction losses have been proposed to address this \cite{lewis2021base,fedus2021switch}. 
Alternatively,~\cite{wang2022image} assigned specific experts for vision, language, and vision-language. 
Here, we introduce a dataset-aware selector to tackle the imbalance issue.
\cite{gao2022parameter} proposed a matrix product operator to reconstruct the matrix in the expert layer and increase model capacity.
In contrast, our work achieves enhancing parameter efficiency by sharing the general summarization ability.
\textcolor{black}{In the domain of summarization, ~\cite{ravaut2022summareranker} proposed one of the few works that related to MoE, where they used an MoE architecture to rerank summary candidates, which is different from our work.}

\textbf{Lightweight Fine-tuning.} 
Lightweight fine-tuning freezes most of the pretrained parameters and modifies the pretrained model with small trainable modules.
Finding high-performing module architectures and the subset of pretrained parameters to fine-tune is the main challenge.
A series of research considers removing parameters by ablating away some model weights \cite{zhao2020masking,radiya2020fine}.
Another line of research considers inserting parameters.
For example, \cite{zhang2020side} trained a side network that is fused with the pretrained model via summation.
Following this direction, \cite{li2021prefix} optimized a small continuous task-specific vector prefixed to the original inputs.
In this work, we design a multi-role mixture of expert structure, where the deputy expert-related are naturally fit for low-resource finetuning.

\section{Background}
We base our summarization model on the prevalent Transformer architecture \cite{vaswani2017attention}, comprised of an encoder and decoder, each with repeated Transformer blocks. 
Each block has a multi-head self-attention sub-layer and a two-layer feed-forward neural network (FFN).
Suppose the self-attention output is $\mathbf{A}$. 
Then, the FFN outputs $\mathbf{X}$ by:
\begin{align}
    \mathbf{H}=\sigma\left(\mathbf{A} \mathbf{W}_1+\mathbf{b}_1\right),\; \; \mathbf{X}=\mathbf{H}\mathbf{W}_2 +\mathbf{b}_2,
    \label{ffn}
\end{align}
where $\mathbf{W}_1 \in \mathbb{R}^{d \times d_h}, \mathbf{W}_2 \in \mathbb{R}^{d_h \times d}, \mathbf{b}_1 \in \mathbb{R}^{d_h}$ and $\mathbf{b}_2 \in \mathbb{R}^d$ are weights of the FFN,  $\sigma$ is the activation function,   $d$ is the embedding dimension, and $d_h$ is the hidden dimension of the FFN.

Mixture-of-Experts was  firstly proposed to facilitate conditional computation and increase the parameter count without altering   the floating point operations for each input~\cite{shazeeroutrageously}. 
Essentially, MoE models consist of multiple expert layers similar to the Transformer layers. 
Each of these layers contains a self-attention mechanism and multiple FFNs (Eq.~\ref{ffn}) in parallel, namely ``experts'', denoted as $\{E_i\}^N_{i=1}$.
Each expert has its own set of learnable weights. 
To keep the computational cost constant, a gating network $G$  outputs a sparse $N$-dimensional vector to route each token via a few experts.

Similar to Eq.~\ref{ffn}, we denote the output of the attention mechanism as $\mathbf{A}$. 
For each $\mathbf{a}_s$ (the $s$-th row of  $\mathbf{A}$) that corresponds to the $s$-th input token, the corresponding output $\mathbf{x}_s$ of FFNs is:
\begin{align}
    \mathbf{x}_s=\textstyle \sum_{i \in \mathcal{T}} G_i\left(\mathbf{a}_s\right) E_i\left(\mathbf{a}_s\right).
    \label{moelayer}
\end{align}
Here, $\mathcal{T} \subset\{1 \cdots N\}$ is the activated set of experts that have the largest $G_i$ values, and $G_i\left(\mathbf{a}_s\right)$ denotes the probability of selecting expert $E_i$.

Various approaches have been proposed to compute $G_i$ and construct $\mathcal{T}$.
A classic method, proposed by~\cite{shazeeroutrageously}, calculates $G_i$ by a weighted matrix $\mathbf{W}$ based on the input $\mathbf{a}_s$:
\begin{align}
G_i\left(\mathbf{a}_s\right)=\left[\operatorname{softmax}\left(\mathbf{a}_s \mathbf{W}\right)\right]_i,
\end{align}
where $\mathbf{W} \in \mathbb{R}^{d \times N}$. 
This method, however, has two major drawbacks:
(1) It often leads to a load imbalance problem, where $\mathbf{W}$ collapses, causing nearly all the inputs to be routed to the same expert \cite{fedus2021switch,zuo2022moebert}.
(2) The gating function lacks awareness of the input dataset's diversity, an important source of information that reflects the attributes of the inputs.

\begin{figure*}[t]
\centering
\includegraphics[scale=0.5]{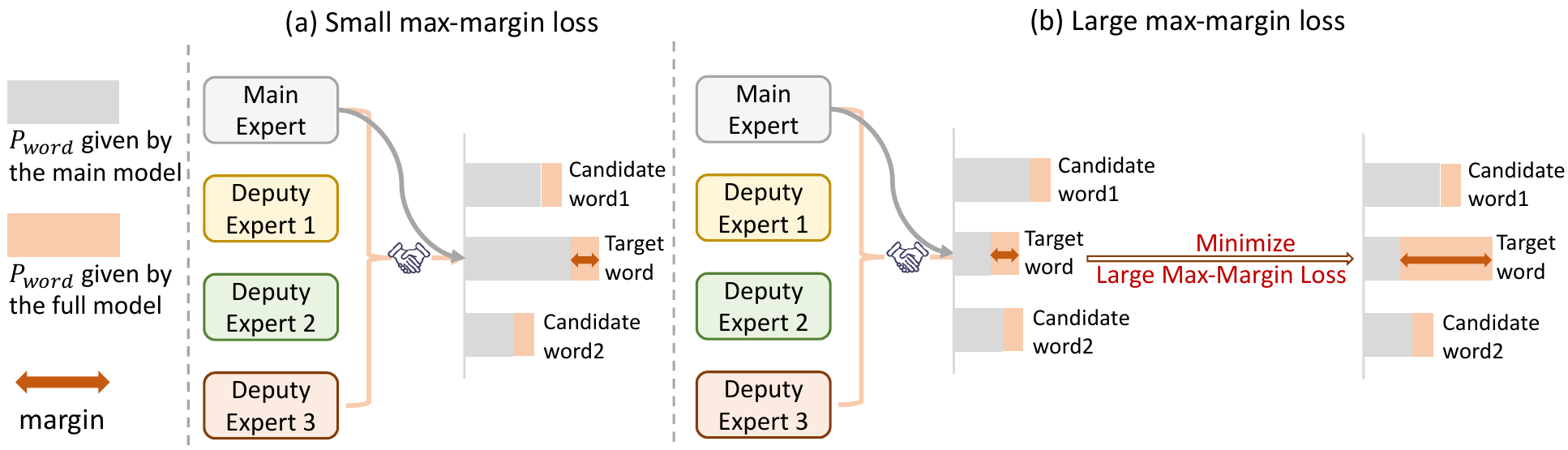}
\caption{
    Examples illustrating  the max-margin loss  $\mathcal{L}_{m}$ in two scenarios. (a) $\mathcal{L}_{m}$ is small when the main expert performs well, where both $P^{\text{full}}_{\text{word}}$ and $P^{\text{main}}_{\text{word}}$ for the target word surpassing other candidates.
    (b) $\mathcal{L}_{m}$ is large  when the main  model cannot perform well. 
    In this scenario, minimizing the  max-margin loss can maximize the margin $m_t$, thus preventing the overconfidence of the main model and stimulating deputy experts to learn to  predict the correct target word.
}
\label{fig:margin}
\end{figure*}

\section{The Proposed MoeSumm Model}
In this section, we first present an algorithm that adapts an MoE into our MoeSumm model.
This structure promotes summarization across diverse datasets by integrating experts for distinct data attributes.
Such a structure enables the summarization ability on different datasets by introducing experts for specific data attributes.
Then, we detail how  MoeSumm can be used in out-of-domain few-shot and zero-shot scenarios.

The overall framework of our model is shown in Figure~\ref{fig:frame}(a).
Our model includes a main expert used for all datasets and \textcolor{black}{a dataset-aware expert selector to choose suitable deputy experts. This dataset-aware selection method overcomes the previously mentioned limitations by ensuring that cases with similar attributes are routed to the proper deputy experts based on the dataset information.}

\subsection{Dataset-aware Expert Selector}

Let $N^p$  denote the number of deputy experts, and $\mathbf{a}_{s,e}$   be the token representation in the $s$-th position of the input sequence from dataset $e$ after the attention process.
Let's consider trainable weight matrices $\mathbf{W}_{e} \in \mathbb{R}^{d \times N^p}$ corresponding to each dataset $e$. 
We multiply the input $\mathbf{a}_{s,e}$ with the dataset-specific weight matrix $\mathbf{W}_{e}$ to  incorporate data information in the gating mechanism, yielding  the routing logits:
\begin{align}
    c_e\left(\mathbf{a}_{s, e}\right)=\mathbf{a}_{s, e}  \mathbf{W}_{e},
\end{align}
where $c_e(\mathbf{a}_{s,e}) \in \mathbb{R}^{N^p}$.
To obtain the routing probabilities, we   normalize  the routing logits  using a softmax over the $N^p$ deputy experts.
The gate value for the $i$-th deputy expert is then given as:
\begin{align}
    G_{i,e}\left(\mathbf{a}_{s, e}\right)=\text{softmax}[c_e\left(\mathbf{a}_{s, e}\right)]_i.
\end{align}

We can now select the top-$k$ gate values for routing the token. 
Following previous works \cite{gupta2022sparsely,zuo2022moebert}, we constrain the gating method to route each token to only the    top-$1$ expert FFN:
\begin{align*}
p=\text{argmax}_iG_{i,e}(\mathbf{a}_{s,e}),  \; \;
    g_p=G_{p,e}(\mathbf{a}_{s,e}),
\end{align*}
where $g_p$ is the highest score.
Following Eq.~\ref{ffn}, we integrate the outputs from main and deputy experts, guided by the gate score:
\begin{align}
    \mathbf{H}&=\sigma\left(\left[\mathbf{A} \mathbf{W}^m_1 +\mathbf{b}^m_1;g_p\left(\mathbf{A} \mathbf{W}^p_1 +\mathbf{b}^p_1\right) \right]\right),\\
    \mathbf{X}&=[\mathbf{W}_2^m;\mathbf{W}_2^p]\mathbf{H}+\mathbf{b}^m_2,
\end{align}
where $[;]$ denotes the concatenation operation, and superscript $m$ and $p$ denote   parameters   from the main and   selected deputy expert, respectively.

In the above formulation, the dataset-aware gating function $\mathbf{W}_{e}$ learns to route input tokens to specialized experts. 
Importantly, the experts don't have a direct relationship with the datasets, but depend on the input context, encouraging information sharing among all experts and datasets.
Note that the experts themselves do not have an explicit relationship with the datasets and are only dependent on the input context, so as to encourage information sharing among all experts and datasets.

\subsection{Max-margin Loss}
The intrinsic difference between our MoeSumm and standard MoE is the roles   assigned  to experts. 
MoeSumm features a main expert that acquires a generalized summarization skill adaptable to diverse datasets, and deputy experts that specialize in handling cases with specific attributes.
Given the difficulty of defining general and specialized summarization targets, we propose a max-margin loss. 
This strategy aims to prevent the model from over-relying on the main expert, thereby ensuring the contributions of deputy experts aren't overshadowed.

As illustrated in Figure~\ref{fig:margin}, we first define the margin as the difference between the predicted probabilities of the full model (with main and deputy experts) and  the main model (using only main expert):
\begin{align}
    m_t=P^{\text{full}}_t\left(y_{t}\right)-P_t^{\text{main}}\left(y_{t} \right),
    \label{mt}
\end{align}
where $y_t$ is the $t$-th token in the summary, and $P^{\text{full}}_t$ and $P^{\text{main}}_t$ denote the predicted probability of the $t$-th token by the full model and the main model, respectively.
Intuitively, a large $m_t$ suggests that the full model significantly outperforms the main model, \textcolor{black}{ highlighting the valuable contributions of deputy experts and the effective collaboration between main and deputy experts.}
if $m_t$ is large, then the full model is apparently better than the main model.
If $m_t$ is small, there are two possibilities.
One is that both the full and the main models perform well, resulting in similar predicted probabilities (both $ P^{\text{full}}_t$ and $P_t^{\text{main}}$ are high).
The other possibility is that the main expert is not good enough but overconfident, thus, leading to subpar performance of both the full and main models (both $ P^{\text{full}}_t$ and $P_t^{\text{main}}$ are low).

Hence, we present the max-margin loss $\mathcal{L}_m$,
which adds a coefficient to the margin:
\begin{align}
    \mathcal{L}_{m}=\textstyle \sum_{t=1}^{n_y}\left(1-P_t^{\text{full}}\right) \left(1-m_t^{5}\right) / 2,
\end{align}
where we abbreviate $P^{\text{full}}_t(y_t)$ as $P_t^{\text{full}}$.
The term $ (1-m_t^{5}) / 2$ is a  monotonically decreasing non-linear function  with respect to $m_t$, which ensures that \textcolor{black}{the minimization of $  \mathcal{L}_{m}$ maximizes $m_t$.} 
We choose a Quintic function (fifth power) here as it offers more stability \cite{miao2021prevent}, which is confirmed in our preliminary experiments.
The first factor in the above equation, $(1-P^{\text{full}}_t)$ accounts for the two scenarios \textcolor{black}, {as illustrated in Figure \ref{fig:margin}}.
When $P^{\text{full}}_t$ is high, the summarization model performs well, requiring minimal optimization on $m_t$.
This is reflected by $(1-P^{\text{full}}_t)$, which acts as a small coefficient of $m_t$.
On the other hand, when $P^{\text{full}}_t$ is low, a large coefficient $(1-P^{\text{full}}_t)$ encourages the maximization of $m_t$, so that the correct target word can be predicted with the help of deputy experts.
The overall loss function of MoeSumm is a combination of text generation loss and max-margin loss.

\subsection{Adaptability of MoeSumm}
Due to its inherent separation of general and specialization ability, MoeSumm has the adaptability to handle few-shot and zero-shot summarization scenarios for out-of-domain data.
Firstly, we can reuse the main expert and only fine-tune the deputy experts along with the expert selector to quickly adapt MoeSumm to a low-resource dataset, as shown in Figure~\ref{fig:frame}(b).
Moreover, in a zero-shot scenario where no training data is available, we can rely solely on the main expert without deputy experts to generate summaries.
This approach is possible as the main expert is competent at producing general summaries, especially when the model lacks prior knowledge about the target domain.

\section{Experiments}
\label{sec:experiment}

\subsection{Dataset and Evaluation setting} 
In the standard evaluation setting,  MoeSumm is trained on a \textit{Mix dataset} comprising of three widely-used summarization datasets: CNN/DM \cite{hermann2015teaching}, WikiHow \cite{koupaee2018wikihow}, and PubMed \cite{cohan2018discourse}, selected for their diverse domains and varying source and target lengths. 
\textcolor{black}{For out-of-domain few-shot evaluation,  MoeSumm is fine-tuned as shown in Figure \ref{fig:frame}(b).  
We use a small number of samples from  XSum~\cite{narayan2018don}, AESLC \cite{zhang2019email}, and Reddit \cite{kim2019abstractive} for fine-tuning the expert selector and deputy experts, and assess the  fine-tuned MoeSumm on the corresponding testing samples.
For zero-shot  evaluation, there is no fine-tuning, the full-scale  MoeSumm trained on Mix dataset is tested on unseen datasets including  Gigaword \cite{napoles2012annotated}, BillSum~\cite{kornilova2019billsum}, arXiv~\cite{cohan2018discourse}, BIGPATENT~\cite{sharma2019bigpatent}, and MultiNews \cite{fabbri2019multi}.}

\renewcommand\arraystretch{1.2}
\begin{table*}[t]
\small
\caption{Performance in in-domain and out-of-domain scenarios.
\textbf{Bold} numbers indicate statistically \textit{significant} improvements over the FlatMoe, using a two-tailed paired t-test
~\cite{dror2018hitchhiker}  at a significance level of 0.05. 
(+\%) is  the average percentage improvement in ROUGE over FlatMoe.
}
\centering
\begin{tabular}{ccccc}
\toprule
\multirow{2}{*}{Dataset} & BART-mix  
 & BART-indiv  &  FlatMoe & MoeSumm  \\
&$R_1$ / $R_2$ / $R_L$ / BS   & $R_1$ / $R_2$ / $R_L$ / BS & $R_1$ / $R_2$ / $R_L$ / BS & $R_1$ / $R_2$ / $R_L$ / BS / (+\%)\\
 \midrule
 \multicolumn{3}{l}{\textit{Trained on Mix dataset, in-domain test}:}&&\\
CNN/DM  &  43.84/20.78/40.53/89.01  & 44.16/21.28/40.90/88.16  &44.09/21.03/40.95/89.02 & \textbf{45.01/21.75/41.91/89.26} (+2\%)  \\
PubMed  & 44.52/17.96/39.62/86.26  & 44.57/17.96/39.70/86.25& 44.74/17.86/39.86/86.38 & \textbf{45.40/18.44/40.54/86.60} (+2\%)\\
WikiHow   & 46.49/20.44/44.90/90.68 & 46.96/20.93/45.41/90.81 & 46.83/20.34/45.05/90.71  & 46.75/\textbf{21.35}/\textbf{45.33}/\textbf{90.89} (+2\%)\\
\midrule
 \multicolumn{3}{l}{\textit{Trained on Mix dataset, out-of-domain zero-shot test}:}&&\\
Gigaword  & 26.21/8.94/23.11/86.04  & 25.14/8.52/22.42/85.45& 25.80/9.06/22.65/85.95 & \textbf{26.57/9.32/23.89/85.87} (+4\%) \\
BillSum  & 43.16/18.78/36.36/84.22  & 41.19/18.04/34.47/83.98& 42.84/18.95/35.74/84.12 & \textbf{43.65/19.22/36.45}/\textbf{84.16} (+2\%)\\
arXiv & 41.41/14.18/36.53/85.39  & 39.58/13.14/33.70/84.02 & 42.05/14.46/37.26/84.66 & \textbf{43.91/15.51/38.55/85.60} (+5\%)\\
BIGPATENT   & 34.82/10.17/29.15/83.78 & 32.55/8.95/27.59/83.82& 35.05/10.37/29.13/83.66 & \textbf{37.02/11.10/31.01/84.18} (+5\%)\\
MultiNews  & 28.69/9.44/25.65/85.25& 27.86/9.34/25.17/83.71 & 28.97/9.73/26.12/85.02 & \textbf{31.62/10.49/28.37/85.48} (+8\%) \\
 
\toprule
 \multicolumn{4}{l}{\textit{Fine-tuned for out-of-domain few-shot test}:}&\\
XSum10  & 32.21/9.01/23.74/88.76& 31.81/8.82/23.33/88.68 & 32.65/9.06/23.87/88.76 & \textbf{33.15/10.22/24.42/89.21} (+5\%)  \\ 
 XSum100 & 35.17/12.05/27.52/89.74  & 34.69/11.77/27.36/89.67 & 35.25/12.03/27.87/89.89 & \textbf{35.58/13.06/28.06/89.94} (+3\%)
\\
\midrule
AESLC10   &  26.17/12.72/23.39/84.38  & 26.56/12.81/23.81/84.18& 26.47/12.83/23.76/84.48 & \textbf{27.48/14.32/25.53/86.04} (+7\%)\\ 
 AESLC100  & 30.44/17.20/28.64/84.98 & 30.01/15.26/26.98/84.94& 31.12/17.34/29.39/86.03 & \textbf{32.87/17.96/30.92/86.54} (+5\%)

\\
\midrule
Reddit10  & 19.39/6.47/17.44/86.97 & 17.56/5.58/15.74/85.45& 20.03/6.89/18.89/87.22 & \textbf{21.74/8.00/20.75/88.09} (+11\%)  \\ 
 Reddit100 & 21.62/8.37/20.45/87.94 &  19.44/7.19/17.34/86.00 & 23.31/9.93/22.59/88.13 & \textbf{25.57/11.45/24.34/88.67} (+10\%)
\\
\bottomrule
\end{tabular}
\label{tb:low_resource_performance}

\end{table*}

\subsection{Baselines}
We first compare MoeSumm with the classic BART \cite{lewis2020bart}, which is a well-known pretraining sequence-to-sequence model.
\textcolor{black}{We assess BART’s performance in two different settings. \textbf{BART-mix}: training a single model on the entire Mix dataset, and  \textbf{BART-indiv}: training separate models on individual datasets within the Mix dataset.}
For BART-indiv, we utilize its CNN/DM version for zero-shot evaluation and fine-tune it in the few-shot scenario. 
The CNN/DM version is chosen due to its superior performance.
We further employ a naive flat \textbf{FlatMoe} baseline, where there is no main expert, and expert selection has no dataset information.
The obtained models after mixed training are directly used for the zero-shot test.

We also show the superiority of our model compared with a prompt-tuning approach Prefix \cite{li2021prefix}, an adapter-based baseline \cite{huh2022lightweight}, and \textbf{GPT-3.5}.
\textbf{Prefix} \cite{li2021prefix} is a prompt-tuning approach that keeps  BART frozen and optimizes a sequence of continuous task-specific vectors appended to the original tokens, denoted as prefix.
\textbf{Light} \cite{huh2022lightweight} is a lightweight meta-learning adapter inserted into the attention mechanism of BART, which is designed for low-resource scenarios.

\subsection{Implementation Details}
We implemented our experiments in Huggingface on 4 NVIDIA A100 GPUs.
We used the BART-large as the pretrained language model by default. 
The expert dimension $d_h$ is 512 and the deputy expert number is 3 by default, \textcolor{black}{for balancing between model complexity and performance.
}.
We used Adam optimizer with $\epsilon$ as 1e-8 and $\beta$ as (0.9, 0.999). 
The learning rate is set to 3e-5. 
The warm-up is set to 500 steps for all experiments. 
The batch size is set to 8 with gradient accumulation steps of 4. 
The encoded length is 1024, and the maximum decode length is 300.
When decoding a summary, we used beam search with a beam size of 4, and the vocabulary size of the model is 50,625. 

For baseline Prefix, we use the code provided by the authors~\cite{li2021prefix}\footnote{\url{https://github.com/XiangLi1999/PrefixTuning}}.
The performance of Prefix on XSum is slightly different from the originally reported result~\cite{li2021prefix}.
Similar observations have been found here\footnote{\url{https://github.com/XiangLi1999/PrefixTuning/issues/2}}.
Nonetheless, Prefix is still a reasonably evaluated baseline.
We use our own implementation of Light, as the original work~\cite{huh2022lightweight} did not provide the code. 
Notably, our version of Light outperforms the reported results in certain metrics, reaffirming the credibility of our reimplementation for evaluation comparisons.

\subsection{Evaluation Metrics}
We employ standard ROUGE F1 ~\cite{lin2004rouge}:  
ROUGE-1 ($R_1$), ROUGE-2 ($R_2$), and ROUGE-L ($R_L$), each indicating the matches of unigram, bigrams, and the longest common subsequence, respectively.
We also use BERTScore (BS) \cite{zhang2019bertscore} to calculate semantic similarities between the summaries. 
Beyond these automatic evaluation metrics, we assess system performance by human judgments.

\subsection{Main Experimental Results} 
\label{main_result}
\textbf{Performance on In-domain Test. }
The first three rows in Table~\ref{tb:low_resource_performance} show the performance of baseline models and our model on in-domain test.
We first observe that BART-indiv performs better than BART-mix in most metrics.
This is expected as the three datasets have distinct attributes that can confuse a mixed single model.
Secondly, FlatMoe performs comparably to BART-indiv and outperforms BART-mix due to its expert structure.
Finally, our MoeSumm model leverages the three datasets more effectively, achieving significantly better results across four metrics on all three datasets.
Specifically, it outperforms  BART-mix by 3\%/5\%/3\% RG-1/RG-2/RG-L scores on CNN/DM respectively.
This demonstrates that combining training datasets can be regarded as a data augmentation method for our MoeSumm with the help of mixture-of-expert structure, though they come from different domains.
This shows that our MoeSumm model, with its hierarchical mixture-of-expert structure, can effectively utilize combined training datasets from different domains as a method of data augmentation.

\textbf{Performance in Out-of-domain Zero-shot Scenarios.}
The adaptability of   models on unseen tasks is  reported in the second block in Table~\ref{tb:low_resource_performance}. 
Models are tested on Gigaword, BillSum, arXiv, BIGPATENT, and MultiNews datasets, which encompass various fields such as news, academic papers, patents, and bills.
BART-mix outperforms BART-indiv, highlighting the benefits of multi-dataset learning for adaptability.
FlatMoe does not show significant improvement compared with BART-mix, indicating that flat MoE structure cannot improve the generalization ability of the model.
MoeSumm demonstrates significantly superior adaptability, outperforming baselines in all metrics. 
It is worth noting that, in the zero-shot scenario, MoeSumm introduces no extra parameters compared to the basic BART model, as the expert selector and deputy experts are not used.
Specifically, our model outperforms Prefix by 1.64 ROUGE-1 scores on arXiv, and 1.55 ROUGE-L scores on BIGPATENT dataset.

\textbf{Performance in Out-of-domain Few-shot Scenarios.}
The third section in Table~\ref{tb:low_resource_performance} and Table~\ref{tb:light} show results where only 10/100 samples from datasets such as XSum, AESLC, and Reddit (spanning news, email, and post domains) are available for fine-tuning. These results are averaged over 10 independent runs. 
BART-mix significantly outperforms BART-indiv, similar to the zero-shot setting. 
MoeSumm achieves better performance than the strong baseline FlatMoe, demonstrating the effectiveness of our hierarchical expert structure in distinguishing general and specialized summarization abilities across various low-resource scenarios.

\begin{table}[htbp]
\centering
\small
\caption{Human evaluation results of three models, in terms of succinctness, informativeness, and fluency of generated summaries.}
\resizebox{0.6\linewidth}{!}{%
\begin{tabular}{l|ccc}
\toprule
Model          & Succ & Inform & Flu \\ \hline
BART-mix       & 2.37 & 2.34   & 2.07 \\
FlatMoe         & 2.42 & 2.39   & 2.28 \\
MoeSumm        & \textbf{2.56} & \textbf{2.62} &\textbf{2.44} \\
\midrule
GPT-3.5        & 2.33 & 2.65& 2.61 \\
\toprule
\end{tabular}
}
\label{human}
\end{table}

\textbf{Human Evaluation.}
We also conducted a human evaluation of our model to balance the potential bias of automated metrics in assessing summarization quality~\cite{E17-2007}. 
We randomly sampled 50 test instances from the CNN/DM, Gigaword, and XSum datasets. 
Following the methodology proposed by \cite{liu2022end}, but with a threefold larger evaluation scale, we presented three Ph.D. evaluators with an article and its corresponding system-generated summaries. 
They were asked to rate these summaries based on Succinctness, Informativeness, and Fluency. 
The score ranges from one to three, where three is the best.
The averaged results are shown in Table~\ref{human}. 
Our model outperforms the baseline models BART-mix and FlatMoe in all metrics. 
Specifically, MoeSumm outperforms BART-mix by 0.19, 0.28, and 0.37 in terms of Succ, Inform, and Flu scores, respectively. 
 We obtain a p-value of $6 \times 10^{-3}$, $2 \times 10^{-7}$, and $9 \times 5^{-5}$ for Succ, Inform, and Flu, respectively.

\subsection{Comparison with GPT-3.5} 
Our human evaluation also consists of a comparison with GPT-3.5.
As shown in Table~\ref{human}, MoeSumm displays superior succinctness and comparable informativeness to GPT-3.5, while GPT-3.5 gives more fluent text.
In evaluation, we also found that MoeSumm produces more concise summaries, whereas GPT-3.5 outputs more conjunctions such as `while' and `although'. 
Furthermore, GPT-3.5 often produces inferred sentences that enhance comprehensibility at the expense of brevity. 
This aligns with previous findings \cite{yang2023exploring} that ChatGPT generally opts for more extended summaries.
Moreover, we conduct a ROUGE comparison between GPT-3.5 and MoeSumm.
Our model significantly outperforms in in-domain datasets and is comparable in out-of-domain and few-shot scenarios.
Taking into account that GPT-3.5 boasts 300 times more parameters, coupled with the recent insights \cite{liu2022revisiting,zhang2023benchmarking} regarding the alignment of the ROUGE metric with human annotations, MoeSumm's performance is commendable.
We anticipate scaling up our model in the future and allocating more resources to it, allowing for a comprehensive comparison with larger language models.

\begin{table*}[htb]
\centering
\small
\caption{Performance of baselines and our model in low-resource training scenarios.
(+\%/+\%) is the average percentage improvement in ROUGE over Prefix and Light.}
\begin{tabular}{cccc}
\toprule
\multirow{2}{*}{Dataset} & Prefix & Light & MoeSumm \\
 & $R_1$ / $R_2$ / $R_L$ / BS & $R_1$ / $R_2$ / $R_L$ / BS & $R_1$ / $R_2$ / $R_L$ / BS \\
\toprule
XSum10 & 32.50/9.84/23.95/88.93 & 32.29/10.14/24.24/89.03 & \textbf{33.15/10.22/24.42/89.21} (+3\%/+2\%) \\
XSum100 & 35.20/12.74/27.57/89.73 & 35.39/12.90/27.83/89.80 & \textbf{35.58/13.06/28.06/89.94} (+2\%/+1\%) \\
\midrule
AESLC10 & 26.45/13.08/24.26/84.84 & 26.59/13.42/24.53/85.06 & \textbf{27.48/14.32/25.53/86.04} (+6\%/+5\%) \\
AESLC100 & 31.58/16.83/29.11/85.87 & 32.02/17.64/29.64/85.85 & \textbf{32.87/17.96/30.92/86.54} (+6\%/+3\%) \\
\midrule
Reddit10 & 19.95/6.88/17.89/87.03 & 20.24/7.24/18.60/87.47 & \textbf{21.74/8.00/20.75/88.09} (+14\%/+10\%) \\
Reddit100 & 23.64/9.89/22.53/87.62 & 24.47/10.34/23.06/88.03 & \textbf{25.57/11.45/24.34/88.67} (+11\%/+7\%) \\
\bottomrule
\end{tabular}
\label{tb:light}
\end{table*}

\section{ ANALYSIS AND DISCUSSION } 
\subsection{Ablation Study}
We removed  the dataset information in the expert selector and max-margin loss to evaluate their impact on MoeSumm during in-domain test, out-of-domain few-shot and  zero-shot test.  
When  the dataset information was removed, the deputy experts were selected only based on the input content.
As shown in Table~\ref{tb:ablation}, this leads to a notable performance drop in all test scenarios, underscoring the importance of introducing our dataset-aware selection.
Additionally, eliminating the max-margin loss resulted in a 4\% ROUGE-2 score reduction in zero-shot and few-shot settings, indicating its role in distinguishing the functions of main and deputy experts.

\renewcommand\arraystretch{1.2}
\begin{table*}[t]
\small
\caption{Ablation study of MoeSumm when Dataset Information (DI) in expert selector and max-margin loss ($\mathcal{L}_m$) are removed. 
\textbf{Bold} numbers indicate  significant improvements over the second-best.
(+\%/+\%) is the average percentage improvement in ROUGE over w/o DI and w/o $\mathcal{L}_m$.
}
\centering
\small
\begin{tabular}{ccccc}
\toprule
\multirow{2}{*}{Test Dataset} &{Test } & MoeSumm w/o DI  &  MoeSumm w/o $\mathcal{L}_m$& MoeSumm  \\
&$R_1$ / $R_2$ / $R_L$ / BS   & $R_1$ / $R_2$ / $R_L$ / BS & $R_1$ / $R_2$ / $R_L$ / BS & $R_1$ / $R_2$ / $R_L$ / BS / (+\%)\\
 \midrule
 CNN/DM & in-domain & 44.42/21.13/40.67/88.57 &43.91/20.91/40.69/88.83  & \textbf{45.01/21.75/41.91/89.26} (+2\%/+3\%) \\
BIGPATENT &0-shot & 36.60/10.62/30.85/83.93 &  36.38/10.25/30.04/83.59 &\textbf{37.02/11.10/31.01/84.18} (+2\%/+4\%) \\
AESLC &100-shot & 32.72/17.16/30.09/85.04 & 32.46/17.33/30.86/86.10 & \textbf{32.87/17.96/30.92/86.54} (+3\%/+2\%)\\
\bottomrule
\end{tabular}
\label{tb:ablation}
\end{table*}

\subsection{Analysis on Expertise Specialization}

\textbf{Different deputy expert exhibits unique characteristics.} 
We first study this problem from qualitative aspect.
Take deputy expert (DE) \#1 and \#3 for example, we found that DE \#3 excels in generating scholarly summaries, while DE \#1 adeptly describes news events. 
DE \#3 is inclined to generate longer and more complex sentences while DE \#1 usually generates simpler sentences. 
We show two \textit{randomly} selected examples generated by our model using different deputy experts on MultiNews and PubMed datasets in Table~\ref{case}.

\begin{table*}[htb]
\small
\caption{Case study of our model with different deputy experts.}
\label{case}
    \centering
    \begin{tabularx}{\textwidth}{l>{\hsize=0.8\hsize}X>{\hsize=1.2\hsize}X}
    \toprule
       Dataset & MoeSumm with DE\#1 & MoeSumm with DE\#3 \\
        \midrule
        MultiNews & Scott Stevens was fired from his job after his employer discovered he was embezzling money from his company to fund his gambling habit. & He gave his wife instructions to avoid responsibility for his losses and keep her credit intact: she was to deposit a check for \$4,000; move her funds into a new checking account; decline to pay the money he owed the Bellagio casino in Las Vegas; disregard his credit-card debt; file her tax returns; sign up for Social Security survivor benefits; and have him cremated. \\
        \midrule
        PubMed & while no study has examined the influence of anxiety on cognition in patients living with pd by directly comparing groups of pd patients with and without anxiety [author annotation: with no detailed information on the experiments.] & using a cross-sectional design, we compared 17 pd participants with anxiety and thirty-three participants without anxiety on the mini-mental state exam (mmse), the parkinsonism rating scale (prs), and the revised barthel index (rbans). \\
        \bottomrule
    \end{tabularx}
\end{table*}

\begin{figure}[htbp]
\centering
\includegraphics[width=0.33\textwidth]{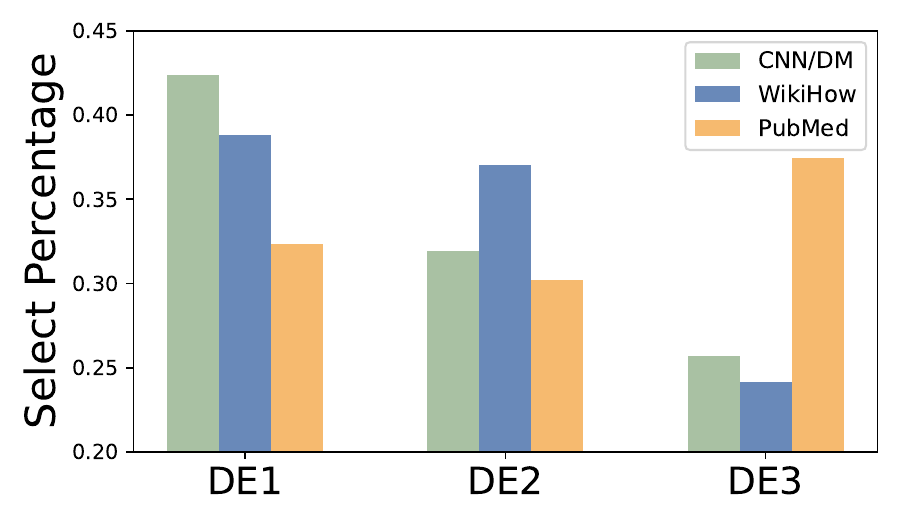}
\caption{Distribution of selected Deputy Experts (DE) associated with three different datasets.}
\label{margin_figure}
\end{figure}

Consequently, we conduct a quantitative analysis. 
MoeSumm with DE \#1 tends to generate shorter sentences (15 words on average), and MoeSumm with DE \#3 can generate longer sentences (37 words on average). 
We also find that with DE \#1, the model obtains a performance of 43.34/16.03/38.29 RG-1/RG-2/RG-L, whereas with DE \#3, the ROUGE performance is improved by 1.4/1.61/1.57 on PubMed.
These observations correspond to Figure~\ref{margin_figure} in our paper, where DE \#1 is more frequently chosen for CNN/DM, and DE \#3 is predominantly selected for PubMed.

\textbf{Deputy experts are utilized differently.}
Second, we assessed how deputy experts are specialized for each dataset.
As illustrated in Figure~\ref{margin_figure}, MoeSumm avoids the pitfall of expert collapse, a situation where inputs are channeled to a single expert~\cite{zuo2022moebert,roller2021hash}. 
Instead, MoeSumm manifests specialized tendencies towards various datasets.
The utilization distribution can also provide an intuitive understanding of the domain-specific abilities each expert acquires. 
For instance, DE \#1 excels at processing news and is therefore predominantly chosen by the CNN/DM dataset, DE \#2 specializes in summarizing user-generated content and is thus frequently utilized by the WikiHow dataset, and DE \#3 is skilled in managing medical information, making it suitable for the PubMed dataset.

\begin{figure}[htbp]
\centering
\includegraphics[width=0.45\textwidth]{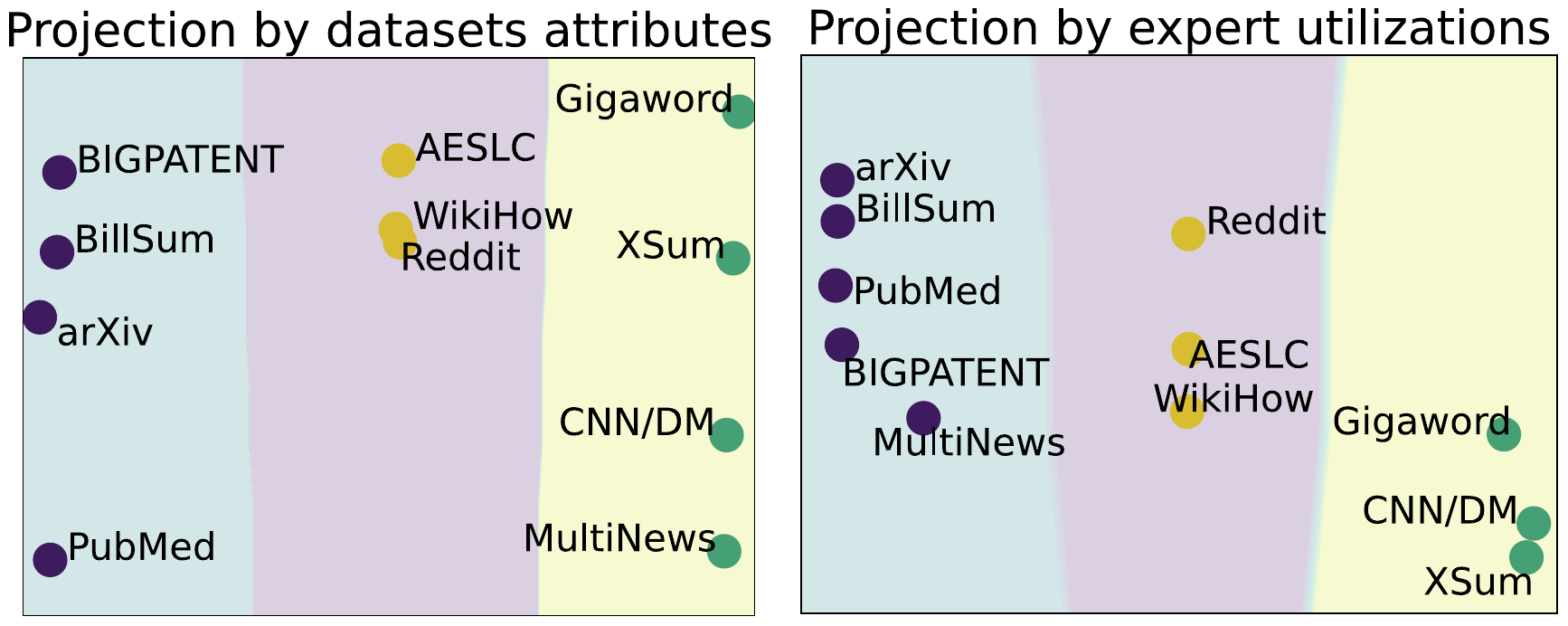}
\caption{Projection comparison between dataset attributes and their deputy expert utilization distribution.}
\label{projection}
\end{figure}
\vspace{-1mm}

\textbf{Deputy expert utilization reflects the dataset attributes.} Finally, we statistically assessed specialized abilities by examining dataset attributes and expert utilization distributions.
If the deputy experts have indeed learned specialized abilities, datasets with similar attributes should select similar deputy experts.
Representing each dataset with a vector [\emph{coverage, density, compression, domain}], where \cite{grusky2018newsroom} defines the first three and \emph{domain} denotes ``news'', ``scholar'', or ``user content'', we mapped each to its deputy expert utilization post MoeSumm fine-tuning on each dataset. 
 Figure~\ref{projection} showcases PCA projection and clustering by attributes and expert utilization. 
It is clear that similar datasets are projected in near space, demonstrating that the MoeSumm learns specialized abilities via the deputy experts.
Notably, MultiNews aligns closely with other  datasets with long documents, indicating that deputy experts are sensitive to document length.

\begin{figure*}[t]
  \centering
  \begin{subfigure}[b]{0.3\textwidth}
    \includegraphics[width=\textwidth]{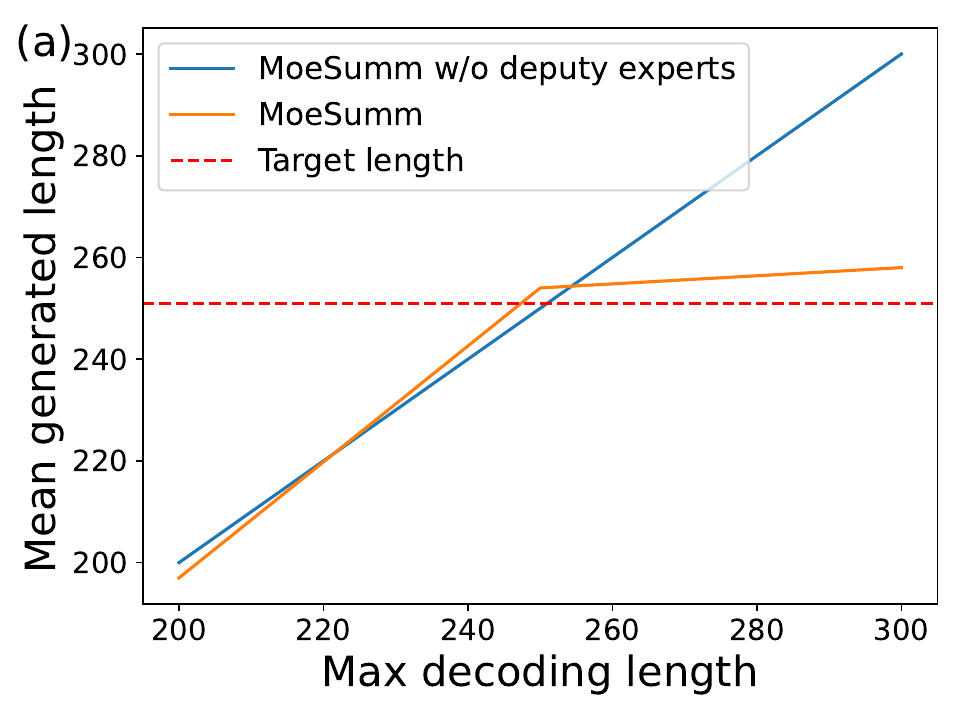}
  \end{subfigure}
  \hfill
  \begin{subfigure}[b]{0.3\textwidth}
    \includegraphics[width=\textwidth]{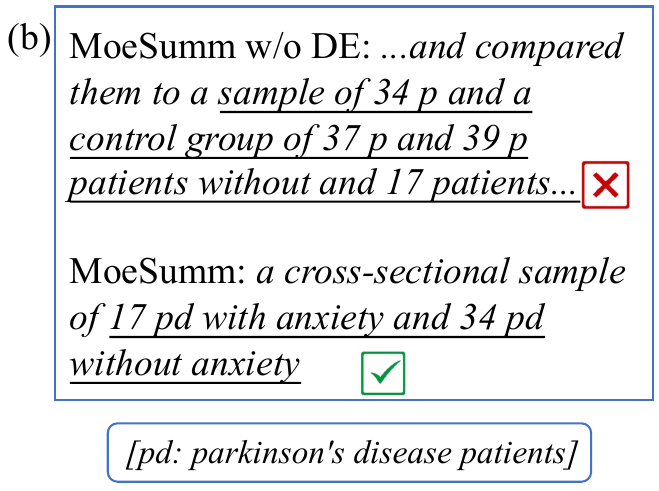}
  \end{subfigure}
  \hfill
  \begin{subfigure}[b]{0.3\textwidth}
    \includegraphics[width=\textwidth]{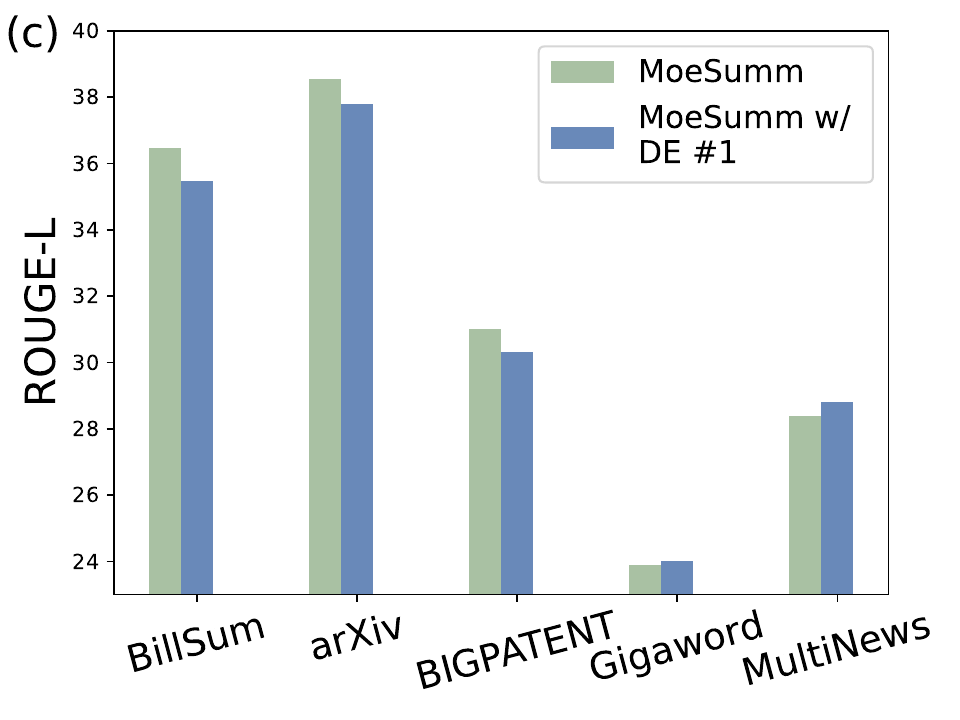}
  \end{subfigure}
  \caption{Analysis of the separation of general and deputy abilities.  (a) Comparing MoeSumm and MoeSumm w/o any deputy experts on the length of the generated  summary. (b) Performance of MoeSumm and MoeSumm w/o deputy experts (DE) for domain-specific context.
  (c) The performance of MoeSumm and MoeSumm with only DE \#1 on five datasets.}
  \label{f2}
\end{figure*}

\textcolor{black}{\textbf{Analysis on General-Specific Expertise Separation.}}
We next investigated whether the general and specialized abilities are indeed separated in MoeSumm.
First, we compare  MoeSumm and MoeSumm w/o any deputy experts (solely the main expert). 
The expectation is that MoeSumm w/o any deputy experts lacks the ability to adapt to target  summary length and language style.  
Figure~\ref{f2}(a),  depicts how the generated summary length varies with the maximum decoding restrictions on PubMed dataset. 
It is evident that MoeSumm w/o deputy experts lacks information regarding the target length, while MoeSumm efficiently halts the generation process to produce an optimized summary length. 
This finding highlights that the deputy experts store domain-specific knowledge.

\begin{figure}[htbp]
\centering
\includegraphics[width=0.2\textwidth]{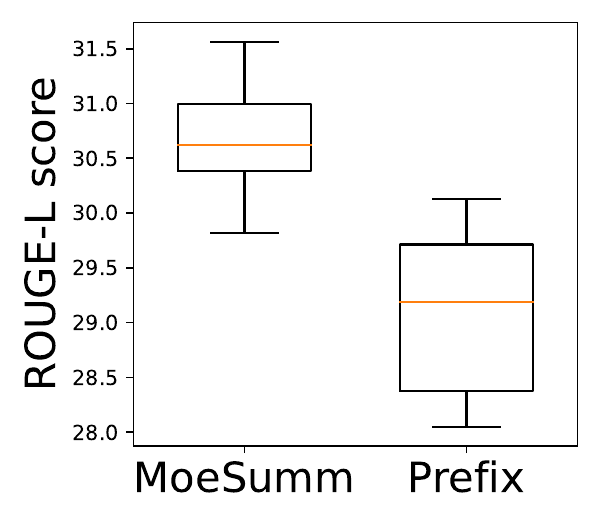}
\caption{Robustness comparison between our MoeSumm and baseline Prefix.}
\label{encoder}
\vspace{-2mm}
\end{figure}

\begin{table*}[htbp]
\centering
\small
\caption{Performance of our framework on other architectures including BART-base and PEGASUS.}
\begin{tabular}{cccccc}
\toprule
Dataset & Setting & BART-base & MoeSumm (BART-base) & PEGASUS & MoeSumm (PEGASUS-version) \\
\hline
CNN/DM & in-domain & 42.45/19.52/39.23/88.52 & \textbf{43.68/20.30/40.37/88.75} (+4\%) & 43.24/20.26/40.08/88.98 & \textbf{44.32/21.13/41.58/89.20} (+2\%)\\
arXiv & 0-shot & 39.56/12.64/34.84/84.86 & \textbf{41.99/13.77/37.05/85.19} (+5\%) & 39.48/13.47/35.54/84.92 & \textbf{41.95/14.48/37.51/85.16} (+4\%)\\
AESLC & 100-shot & 27.74/15.71/26.66/85.88 & \textbf{29.71/17.14/28.54/86.46} (+5\%) & 29.17/16.97/27.39/84.36 & \textbf{30.99/18.10/29.19/85.71} (+5\%)\\
\bottomrule
\end{tabular}
\label{tab:combined_performance}
\end{table*}

Figure~\ref{f2}(b) presents a \textit{randomly sampled} case where the full MoeSumm model and MoeSumm w/o deputy experts summarize a PubMed paper on Parkinson's disease patients, abbreviated as ``pd''. 
The results show that MoeSumm w/o deputy experts struggles to comprehend and accurately employ the key phrase, whereas MoeSumm properly mentions that the experiment was conducted on 17 Parkinson's disease patients with anxiety and 34 patients without anxiety.
This indicates that the deputy experts carry the specialized ability to understand domain-specific terms.

Finally, Figure~\ref{f2}(c) shows the performance of MoeSumm with the main expert and only DE \#1 across five datasets.
According to the analysis from  Figure~\ref{margin_figure}, DE \#1 is proficient at handling news articles.
This is reflected in Figure~\ref{f2}(c). 
Compared to   MoeSumm with all experts,    the model equipped with only DE~\#1 excels in summarizing news domain datasets like Gigaword and MultiNews, but underperforms in other domains like scholarly papers and bills.
This comparison confirms the general summarization proficiency of the main expert, and the flexibility of MoeSumm in selecting suitable deputy experts to complement the main expert, resulting in effective performance across diverse datasets and domains.

\subsection{Efficiency \& Robustness}

We first analyze the parameter \textbf{efficiency} of our model theoretically.
Let's denote the number of experts as $N^p$, the number of layers as $L$, and the number of parameters in each FFN expert as $P_f$. 
Consequently, the total quantity of expert parameters within the model can be calculated as $L \times N^p \times P_f $. 
It's important to note that these experts are shared across all datasets; hence, augmenting the number of datasets does not affect the count of expert parameters.

Conversely, the gating network changes with the dataset, and its parameter count increases with the addition of more training datasets.
If we define $H$ as the dimension of the hidden state and $T$ as the number of datasets, then the quantity of gating parameters can be expressed as $L \times N^p \times H \times T$.
In practical scenarios, the hidden state dimension and the number of datasets are typically far less than the number of FFN parameters, i.e., $H \times T \ll P_f$. 
Hence, the augmentation of training datasets results in a comparatively smaller increase in parameters, especially with the parameters inherent in standard feed-forward Transformer networks.

Next, we discuss about \textbf{robustness}. 
Prior research has suggested that prompt-based fine-tuning might result in high variance~\cite{koksal2022meal}. 
Correspondingly, we evaluated MoeSumm and Prefix using 10 distinct seeds, which led to a variety of training data selections. 
From the box plot in Figure~\ref{encoder}, we can see that MoeSumm generally outperforms Prefix with higher median ROUGE-L scores and a tighter interquartile range, and the range of scores for MoeSumm is also higher, suggesting that even at its worst, MoeSumm performs at a level close to the best of Prefix.
This indicates a more robust performance of MoeSumm.

\textbf{Performance on Other Architectures.}
We also evaluate our architecture using alternative backbones. 
For this purpose, we select BART-base, a smaller-scale model, and PEGASUS, which is specifically designed for summarization tasks and has demonstrated strong performance. 
The comparative performance of these two models, both with and without our mixture-of-experts structure, is illustrated in Table~\ref{tab:combined_performance}.
The consistently superior performance of our framework across various scales and architectures demonstrates that its benefits are not dependent on specific structural design.

\section{Conclusion}

In this paper, we enhanced the flexibility and adaptability of summarization by introducing a parameter-efficient model based on a modified mixture-of-experts structure.  The model consists of a main expert that learns general ability to identify important information, and deputy experts that adapt to domain-specific summary styles.
Our model can be readily applied to diverse summarization datasets and adapted for out-of-domain situations.
Experimental results showed that our model outperforms strong baselines.
In the future, we would like to test the performance of our architecture on larger pretrained language models.

\section*{Acknowledgments}
We would like to thank the anonymous reviewers for their constructive comments. 
The work was supported by King Abdullah University of Science and Technology (KAUST) through grant awards FCC/1/1976-44-01, FCC/1/1976-45-01, REI/1/5234-01-01, RGC/3/4816-01-01, REI/1/5414-01-01, REI/1/5289-01-01, and REI/1/5404-01-01.

\printbibliography

\end{document}